\documentclass[letterpaper]{article} 
\usepackage{aaai2026}
\usepackage{times}  
\usepackage{helvet}  
\usepackage{courier}  
\usepackage[hyphens]{url}  
\usepackage{graphicx} 
\usepackage{natbib}  
\usepackage{caption} 
\usepackage{algorithm}
\usepackage{algorithmic}

\usepackage{newfloat}
\usepackage{listings}
\DeclareCaptionStyle{ruled}{labelfont=normalfont,labelsep=colon,strut=off} 
\floatstyle{ruled}
\newfloat{listing}{tb}{lst}{}
\floatname{listing}{Listing}
\usepackage{booktabs}
\usepackage{amsmath}

\title{
\begin{minipage}{0.9\textwidth}
\centering
    \raisebox{-0.3\height}{\includegraphics[height=1.2cm]{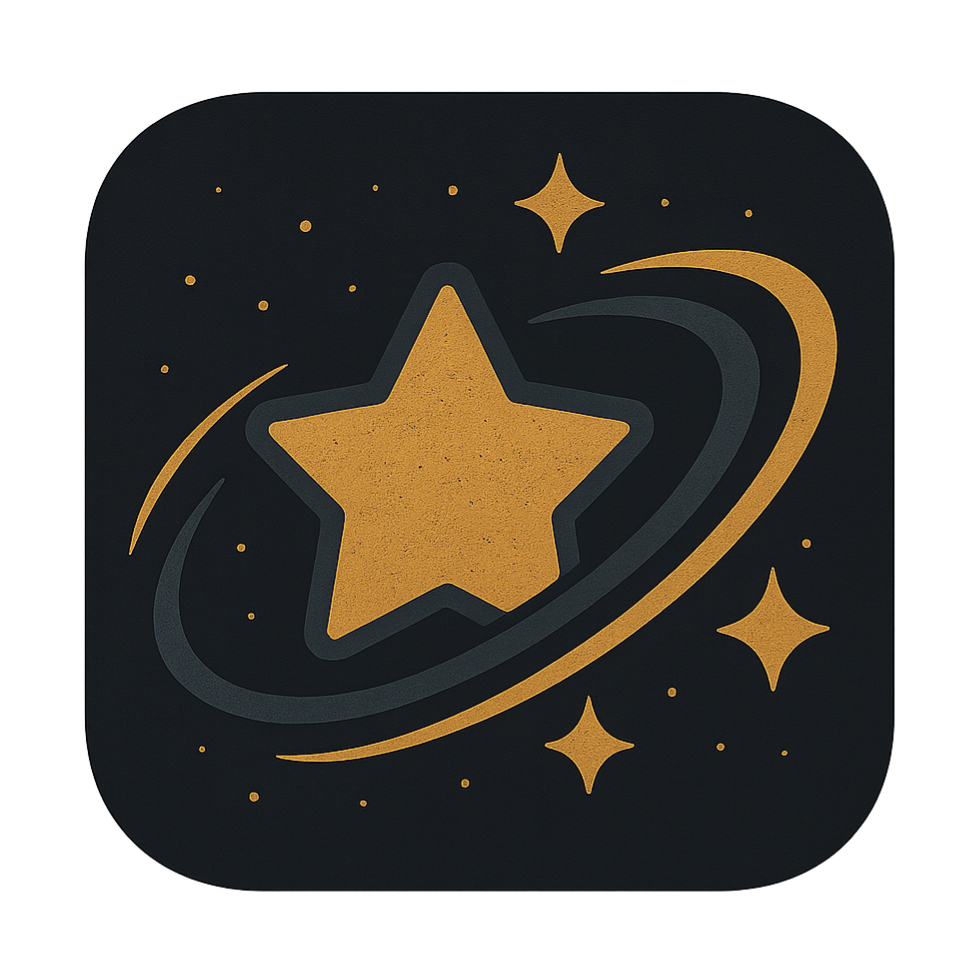}}
    \hspace{0.5em}
    \textbf{Galaxy: A Cognition-Centered Framework for Proactive, Privacy-\\
    Preserving, and Self-Evolving LLM Agents}
\end{minipage}
}
\author {
    Chongyu Bao\textsuperscript{\rm 1}\thanks{Equal contribution}\footnotemark[2],
    Ruimin Dai\textsuperscript{\rm 3}\footnotemark[1],
    Yangbo Shen\textsuperscript{\rm 1}\footnotemark[1]
    Runyang Jian\textsuperscript{\rm 4},\\
    Jinghan Zhang\textsuperscript{\rm 3},
    Xiaolan Liu\textsuperscript{\rm 2},
    Kunpeng Liu\textsuperscript{\rm 3}\thanks{Corresponding author}\\
}
\affiliations {
    \textsuperscript{\rm 1}Carnegie Mellon University,
    \textsuperscript{\rm 2}University of Bristol\\
    \textsuperscript{\rm 3}Clemson University,
    \textsuperscript{\rm 4}Portland State University \\
    chongyub@andrew.cmu.edu, xiaolan.liu@bristol.ac.uk, kunpenl@clemson.edu
}

\usepackage{bibentry}
\nocopyright

\begin{document}

\maketitle

\begin{abstract}
Intelligent personal assistants (IPAs) such as Siri and Google Assistant are designed to enhance human capabilities and perform tasks on behalf of users. The emergence of LLM agents brings new opportunities for the development of IPAs. While responsive capabilities have been widely studied, proactive behaviors remain underexplored. Designing an IPA that is proactive, privacy-preserving, and capable of self evolution remains a significant challenge. Designing such IPAs relies on the cognitive architecture of LLM agents.
This work proposes Cognition Forest, a semantic structure designed to align cognitive modeling with system-level design. We unify cognitive architecture and system design into a self‑reinforcing loop instead of treating them separately. Based on this principle, we present Galaxy, a framework that supports multidimensional interactions and personalized capability generation. Two cooperative agents are implemented based on Galaxy: KoRa, a cognition‑enhanced generative agent that supports both responsive and proactive skills. Kernel, a meta‑cognition‑based meta‑agent that enables Galaxy’s self evolution and privacy preservation.
Experimental results show that Galaxy outperforms multiple state‑of‑the‑art benchmarks. Ablation studies and real‑world interaction cases validate the effectiveness of Galaxy. \footnote{All code is available at https://github.com/Kilo377/GalaxyIPA.}
\end{abstract}


\section{Introduction}
Intelligent personal assistants (IPAs) such as Siri~\cite{apple2025siri} and Alexa~\cite{amazon2023alexa} have been deeply embedded in daily life, improving human abilities to handle complex tasks. Rapid progress of LLM has greatly improved causal reasoning and task-planning in IPAs~\cite{yao2023react, zhang2025nemotronresearchtool, yao2023treethoughts, zhang2024ratt}. LLM-based agents can now understand natural language intent, decompose multi-step plans and invoke tools, which significantly extends their competence~\cite{li2024personal}. Prior work on LLM-based agents generally falls into three categories. Conversational agents~\cite{wahde2022conversational, guan2025evaluatingllmbasedagentsmultiturn, jiang2021automaticevaluationdialogsystems} interact through dialogue and perform tasks by calling external tools. Autonomous agents~\cite{chen2025scalingautonomousagentsautomatic, belle2025agentschangeselfevolvingllm} operate within specific environments and focus on single-task execution. Multi-agent systems~\cite{zhou2024metagpt, chen2024autoos} divide tasks across multiple agents to support collaboration and scalability. Although these systems have achieved great success, they mainly focus on responsive behaviors, but offer limited support for proactive skills that allow agents to act without explicit commands. Addressing this limitation requires overcoming three key challenges. First, proactive behavior requires multi-source perception to support deep user modeling for intent prediction ~\cite{lu2024proactiveagentshiftingllm}. Second, proactive service may raise privacy risks~\cite{hahm2025enhancingllmagentsafety}. Third, LLM agents need the capability to continually adapt their internal architecture and interaction strategies to better support personalization. There are rarely a few works that have explored these aspects jointly. Therefore, the following research question is raised: \textit{Can we design an IPA that is proactive, privacy-preserving, and self-evolving?}

A cognitive framework determines the internal modules of an LLM agent, the observable environment, the actions it can take, and the form of reasoning it can perform~\cite{sumers2023cognitive, toy2024metacognition}. Constrained by their predefined cognitive frameworks, existing LLM agents can only reason within fixed pipelines and do not engage in system‑level design or modification of their own architectures. Some recent studies explore agents that can inspect and revise their own code~\cite{li2023metaagents, liu2025trulyselfimprovingmetacog}, but these efforts often lack integration with task context or system constraints. In practice, cognitive architecture and system design are deeply interdependent, yet they are usually developed in isolation.
Based on these insights, we propose the \textbf{Cognition Forest}, a tree-structured semantic mechanism that connects cognitive modeling with system design. Each subtree of Cognition Forest embeds both agent cognition and metacognition and also integrates design principles and code for reuse by different agents. Based on this, we further develop a system that deeply integrates cognitive architecture with system design, forming a self-reinforcing loop:
\textit{The cognitive architecture drives the system design, and in turn, the improvements in the system design can refine the cognitive architecture itself.}

Guided by this design philosophy, we develop Galaxy, a cognition-enhanced LLM agent framework that supports both responsive and proactive assistance and helps users plan and execute tasks. Galaxy offers multidimensional interaction modalities beyond chat window and can generate or aggregate new cognitive capabilities to support personalized needs. Within Galaxy, we implement two collaborative agents. KoRa is a cognition‑enhanced generative agent that executes tasks via a cognition‑to‑action pipeline grounded in the Cognition Forest, whose long‑horizon semantic constraints help mitigate persona drift in generative agents and improve consistency.
Kernel is a metacognition empowered meta agent that supervises and optimizes the Galaxy framework, self-reflects on capability limitations, expands functionality in response to user demands, and invokes cloud models safely through Privacy Gate. Our work makes three main contributions. 
\begin{itemize}
  \item We propose the Cognition Forest, a semantic structure that integrates cognitive architecture with system design. Based on this foundation, we build Galaxy, an LLM agent framework designed for proactive task execution, privacy-preserving operation, and continuous adaptation.
  \item We implement two collaborative agents: KoRa, a human‑like assistant, and Kernel, a meta agent that maintains stability and drives self-evolution.
  \item Galaxy outperforms multiple state‑of‑the‑art benchmarks. Through ablation studies and case‑study analysis, we validate Galaxy’s effectiveness and provide a real‑world interaction example.
\end{itemize}
We take a step forward of IPAs by arguing that an agent’s understanding of its users should not be constrained by a fixed cognitive architecture, but should evolve through continuous reflection on and refinement of its own system design.

\section{Related Works}
LLM agent based IPA systems can understand natural user instructions and execute complex tasks. For example, LLMPA~\cite{guan2023intelligent} is embedded in mobile applications and completes complex operations under natural language instructions. WebPilot~\cite{zhang2025webpilot} and Mind2web~\cite{deng2023mind2web} are GUI agents that perform multi-step interactions on arbitrary websites. MetaGPT~\cite{zhou2024metagpt} produces efficient solutions through multi-agent collaboration. These works advance the automotive ability of LLM agents. However, current research only gives limited attention to proactive skills, which provide services without explicit instructions. For example,~\cite{liao2023proactive} can infer user intent but remain confined to dialog interactions. Designing proactive agents that directly trigger concrete operations and also address trust and privacy risks in particular when using cloud-based LLM inferences ~\cite{zeng2024privacyrestore} remains a key challenge. In addition, many studies argue that LLM agents should possess self evolution to achieve continuous self improvement~\cite{li2024personal,wang2024survey}. There are rarely research works addressing proactive skills, privacy preservation, and self evolution together. This work tries to take a step forward.

The design of a cognitive architecture directly determines the overall performance of the agent \cite{sumers2023cognitive}. It usually does not include the system design itself. For example, Generative Agents~\cite{park2023generative} use a memory stream, reflection, and planning modules to simulate consistent human-like behavioral patterns. Building on this basis, some work introduces metacognition~\cite{toy2024metacognition}, which enables an agent to examine itself and improve its reasoning process. For example, Metaagent-P~\cite{yanfangzhou2025metagent} improves future performance by reflecting on the current workflow. However, the depth and breadth of such metacognitive ability remain constrained by the underlying cognitive architecture. Further research explores meta agents that examine and improve the system design itself and that automatically generate stronger modules or new agents~\cite{hu2024automated, yin2024g}. Such automated design still relies on preset evaluation standards and struggles to achieve sustained, open-ended evolution. This work implements alternating optimization of system design and cognitive framework. This approach enables a continuously self-evolving system that is guided by user needs.

\begin{figure}[h]
  \centering
  \includegraphics[width=1\linewidth]{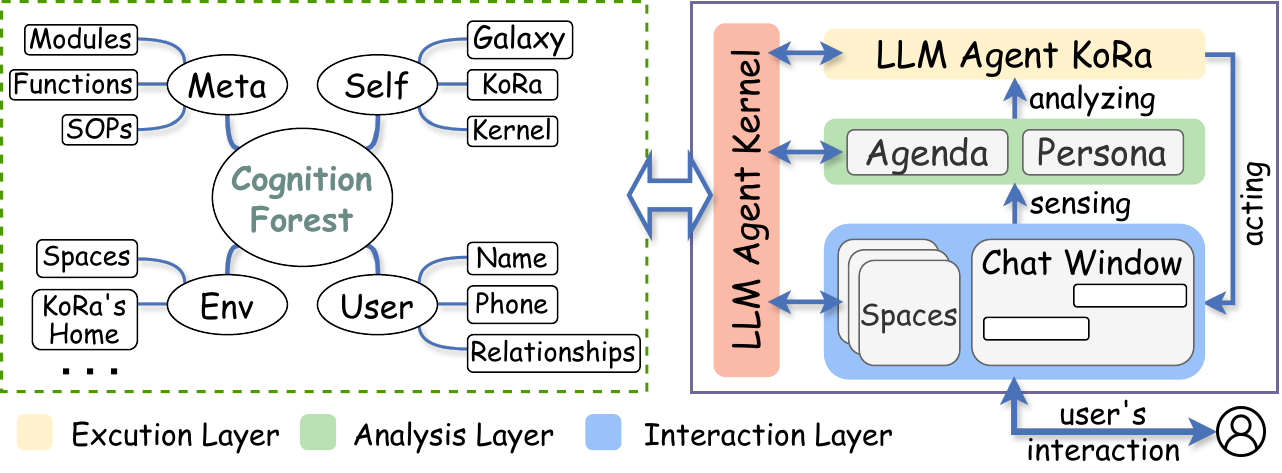}
  \caption{Framework of proposed Galaxy IPA.}
  \label{fig:framework}
\end{figure}

\begin{figure*}[htbp]
  \centering
  \includegraphics[width=1\textwidth]{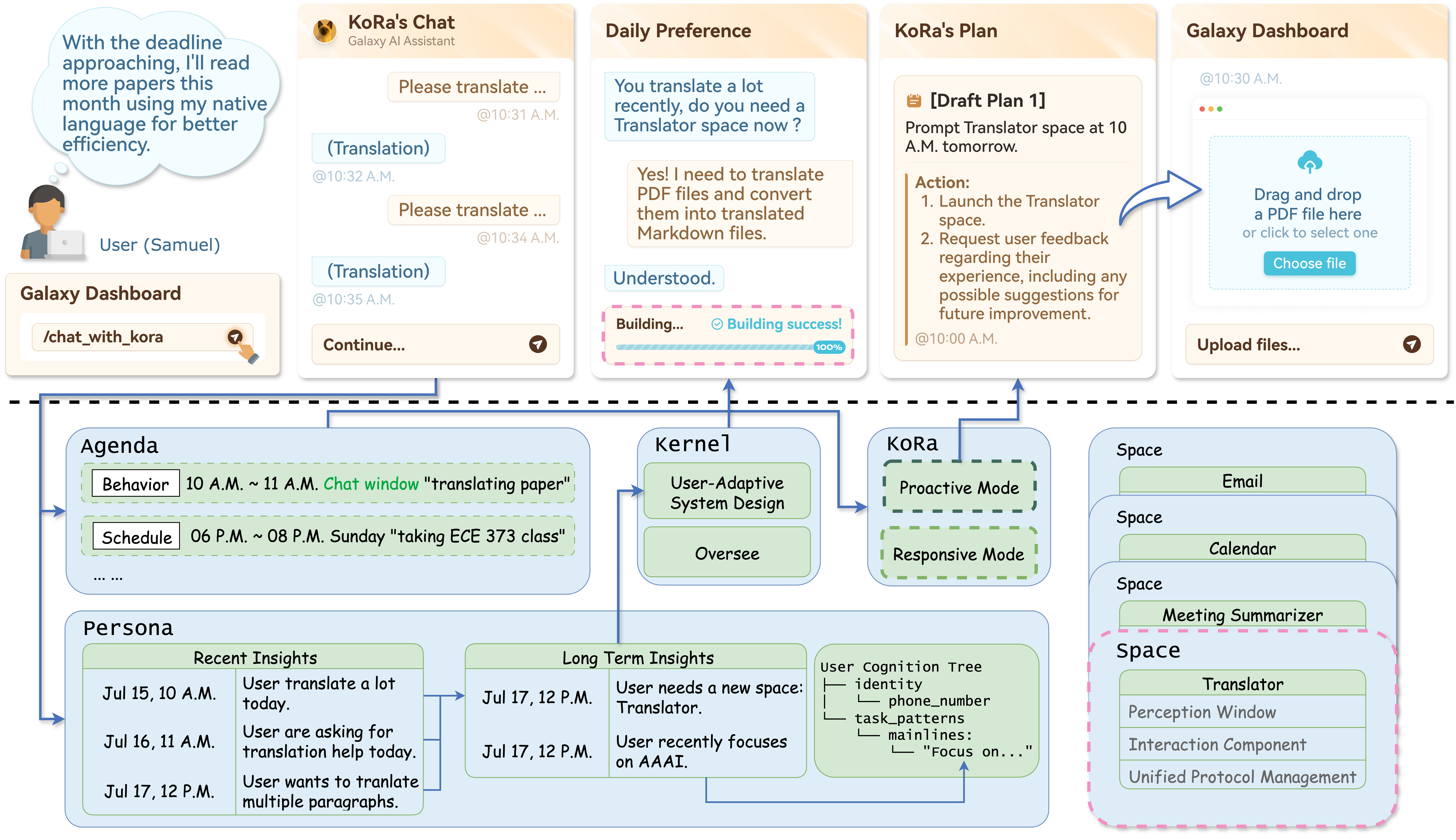}
  \caption{Overall pipeline of proactive skills of Galaxy. \textbf{Upper} part shows the real user interface. \textbf{Lower} part illustrates the corresponding internal execution process, including user modeling, plan generation, and tool activation.}
  \label{fig:main_pipe}
\end{figure*}

\section{Framework}
This section presents an overview of the proposed Galaxy framework, its key components, and the user interaction modalities it supports.

To realize proactive skills, privacy preservation and self evolution, we propose Cognition Forest with Galaxy framework. From Figure \ref{fig:framework}, users interact with Galaxy via Spaces and Chat Window, where the personal assistant KoRa engages in dialogue or autonomously assists in task completion when necessary.
Galaxy is centered on three key components: the Cognition Forest, KoRa, and Kernel. However, realistic operation also depends on supporting components such as interaction modules and analytical units. We systematically introduce the framework following the Perception–Analysis–Execution paradigm. The Interaction Layer perceives the user’s interaction state and contextual signals. The Analysis Layer stores and organizes user-related data and conducts both short-term and long-term user modeling. The Execution Layer generates plans, schedules tasks, and executes actions.
Key modules include: \textbf{Cognition Forest.} The framework’s unified cognitive and metacognitive architecture, structured as multiple semantic subtrees to provide KoRa and Kernel with comprehensive, hierarchical cognitive context.
\textbf{Spaces}. Cognition-driven personalized interaction modules that capture multi-dimensional information during user interactions. \textbf{Agenda.} Models user behavior from perceived event information, generating scheduling recommendations to guide KoRa’s autonomous actions. \textbf{Persona.} Performs comprehensive, long-term modeling of user characteristics to support KoRa’s delegated decision-making. \textbf{KoRa.} A cognition-enhanced generative agent capable of proactive delegation without explicit instructions, or efficient execution when instructed. \textbf{Kernel.} A metacognition-empowered meta agent operating outside the three layers, responsible for maintaining system stability, safeguarding privacy, and enabling Galaxy’s evolution.

\section{Methodology}
In this section, we present a real‑world Galaxy interaction example to highlight the importance of cognitive architecture design, followed by a layer‑by‑layer explanation of each module. Finally, we show how the proposed Cognition Forest enables alternating optimisation of cognitive architecture and system design.

\subsection{A Real Interaction Example}
From Figure \ref{fig:main_pipe}, in recent weeks, during his regular working hours, user Samuel frequently uses the Chat Window to translate academic papers into his native language for reading. This interaction pattern is not convenient. Through behavioral modeling in Analysis Layer, Galaxy identified this as a persistent behavior pattern and recognized that its current capabilities did not fully meet the user’s needs. After further alignment with the user, Galaxy clarified the design requirements and autonomously generated a more convenient translation tool. The system then proactively launches this tool during Samuel’s habitual usage periods to improve his work efficiency. The key to enabling such proactive intelligence lies in the cognitive architecture design: the system must perceive user behavior, analyze user preferences, reflect on its capability boundaries, generate actionable tools, formulate plans, and execute them proactively.

\subsection{Cognition Forest}
We propose the Cognition Forest, which unifies different cognition dimensions and their underlying system designs in a set of tree structures.

\subsubsection{Definition.} Cognition Forest \(\mathcal{F}\) is an structured forest consisting of four subtrees $\mathcal{F} = \{\mathcal{T}_{\text{user}}, \mathcal{T}_{\text{self}}, \mathcal{T}_{\text{env}}, \mathcal{T}_{\text{meta}}\}$,
where \(\mathcal{T}_{\text{user}}\) represents personalized modeling of the user. \(\mathcal{T}_{\text{self}}\) describes the Galaxy itself, its internal agents like KoRa, and their roles and capabilities. \(\mathcal{T}_{\text{env}}\) represents the operational environment including perceivable Space modules, system tools. \(\mathcal{T}_{\text{meta}}\) represents the system’s metacognition like execution pipelines.
In addition to integrating the cognition and metacognition required by a cognitive architecture, the uniqueness of Galaxy lies in associating each cognitive element with its corresponding system design. This means the LLM agents not only know \textit{what to do} and \textit{how to do it}, but also understand \textit{how it is implemented}. This extends the framework’s depth of metacognition beyond the constraints of traditional cognitive architectures.
Each node of subtrees is represented by three dimensions: Semantic, Function, and Design, corresponding to LLM’s semantic understanding, mapped system function, and its concrete implementation. For example, in Memo Space, write\_text node has semantic meaning of writing new content to memo, its functional mapping is the function \texttt{write\_text()}, and its design is the actual implementation code. When a newly added node fails during execution, Kernel can reflect on whether the failure is due to incorrect execution sequence or possible implementation error, and then perform deeper modifications.

\subsection{Sensing and Interaction Protocol}

\subsubsection{Objective.}
Most IPAs' cognitive architectures are constrained by their underlying system design. Although some systems offer limited extensibility at Interaction Layer through mechanisms such as MCP integration~\cite{fei2025mcpzero,lumer2025scalemcp}, existing work rarely extends cognitive depth to the level of concrete interactive functions, thereby limiting personalization capabilities.
\subsubsection{Approach.}
Galaxy addresses this limitation through \textbf{Space}, a protocol that encapsulates heterogeneous information sources into unified modules that are cognitively accessible and interactable. Each Space function is considered as a local execution container and as an independent subtree within Cognition Forest. Spaces can be either customized by users or generated automatically, thereby expanding the system’s perceptual scope and interactive capabilities. Each Space consists of the following components: 
\begin{itemize}
    \item \textbf{Perception Window} continuously observes user actions and environmental signals. It also converts raw inputs into structured \texttt{TimeEvent} entries and state snapshots. These are unified into a consistent, temporally grounded context and delivered upstream to Analysis Layer.
    \item \textbf{Interaction Component} may act as a standalone, personalized module that exposes a user interface and interaction nodes accessible to both the user and KoRa. 
    \item \textbf{Cognitive Protocol} provides a unified development and integration standard for all Spaces. It specifies how high-level intents are translated into concrete system operations. It ensures that each Space can be consistently embedded into the Cognition Forest for further reasoning and task execution.
\end{itemize}
While some LLM agents support automatic function module generation~\cite{wang2023voyager}, Galaxy’s Spaces go further: they are embedded within the system’s cognition and function as integral organs instead of just detachable tools.

\subsection{User Behavior and Cognitive Modeling}
\subsubsection{Objective.}
To support proactive skills, Galaxy anticipates and interprets upcoming events by modeling both explicit schedules and implicit behavioral patterns. It derives stable user cognition from heterogeneous and fragmented natural language signals, constructing a coherent understanding across time and context.
\subsubsection{Approach.} 
Agenda uses a unified \texttt{TimeEvent} to represent two event types. Schedule denotes explicit user schedules (e.g., `class at 18:30 on June 18'). Behavior denotes observed operational actions (e.g., `translated documents in the chat\_window in the morning'). As shown in Figure \ref{fig:main_pipe}, Interaction Layer extracts event content and time ranges and writes them to \textit{Schedule Draft}. Uncertain or conflicting events are routed to an alignment queue for further resolution. All \texttt{TimeEvent} entries are retained to support long-term behavior modeling. Each behavior is represented as a structured triple consisting of time, tool, and semantic intent. Galaxy clusters these behaviors along the tool and semantic dimensions to identify recurring Behavior Patterns. Based on the user’s schedule, Agenda drafts an initial plan and suggests relevant Behavior Patterns for open time slots. The proposed daily plan is then shared with the user for confirmation. Once approved, a summary of next-day actions is passed to KoRa to support timely assistance.


Persona maintains a growing User Cognition Tree $\mathcal{T}_{\text{user}} $, which is organized as a subtree of Cognition Forest $\mathcal{F}$. Galaxy uses LLMs to aggregate dialogues and space interactions into user insights. Each insight contains a natural language summary and a semantic embedding. These insights are high-level semantic cognition rather than statistical aggregates. Similar insights that accumulate within a dimension beyond a threshold are promoted to a long-term node. Insights similar to an existing node are merged and refresh the node timestamp. Nodes that remain unused for a long period decay and are removed. Stable identity information, such as name and phone number, is inserted into the identity branch upon first discovery.

\subsection{KoRa: Intelligent Butler for User}
\subsubsection{Objective.}
Unlike traditional LLM agents, KoRa can proactively manage a user's schedule while also handling real-time requests through dialogue, including immediate tool execution. These two modes follow different triggering mechanisms and may occasionally lead to overlapping actions. For instance, KoRa might pre-schedule a ticket booking for an upcoming trip, while the user later requests the same action manually. In such cases, the system needs to recognize potential duplication and resolve it appropriately. Supporting both proactive and responsive behaviors requires maintaining a consistent task history and avoiding conflicting operations.

\subsubsection{Approach.}
KoRa adopts a generative agent architecture~\cite{park2023generative} with memory stream, planning, and reflection modules to support proactive skills and interactive human-like behavior. To handle interruptions and resume execution in responsive mode, KoRa uses a structured state stack instead of a simple memory stream. The state stack records task type, source, and execution details. KoRa follows a top-down execution flow to advance tasks in the daily plan generated from the \textit{Schedule Draft} provided by Analysis Layer.

\subsubsection{Cognition-Action Pipeline.}
To address issues such as personality forgetting and behavior drift, KoRa introduces an integrated cognitive architecture that connects perception, reasoning, and behavior generation. The foundation of the architecture is Cognition Forest, a hierarchical semantic space that supports intent parsing, semantic routing, and the construction of behavior chains.
KoRa is responsible for both responsive and proactive skills, using cloud-based LLM inference. To ensure privacy isolation, KoRa will maintain only a subset of the Cognition Forest $\mathcal{F}$, referred to as 
$\mathcal{F}^{\text{KoRa}} = \{\mathcal{T}_{\text{user}}, \mathcal{T}_{\text{self}}^{\text{KoRa}}, \mathcal{T}_{\text{env}}^{\text{KoRa}}, \mathcal{T}_{\text{dialogue}}\}$,
where \(\mathcal{T}_{\text{user}}\) is User Cognition Tree maintained by Persona. \(\mathcal{T}_{\text{self}}^{\text{KoRa}}\subset \mathcal{T}_{\text{self}}\) models KoRa’s own capabilities, and responsibilities. \(\mathcal{T}_{\text{env}}^{\text{KoRa}} \subset \mathcal{T}_{\text{env}}\) represents the subset of Galaxy’s environment that KoRa can actively interact with, such as any callable elements within Spaces. \(\mathcal{T}_{\text{dialogue}}\) collects fallback or vague-intent utterances, serving as the default entry point for open-ended interactions.

\begin{figure}[htbp]
  \centering
  \includegraphics[width=1\linewidth]{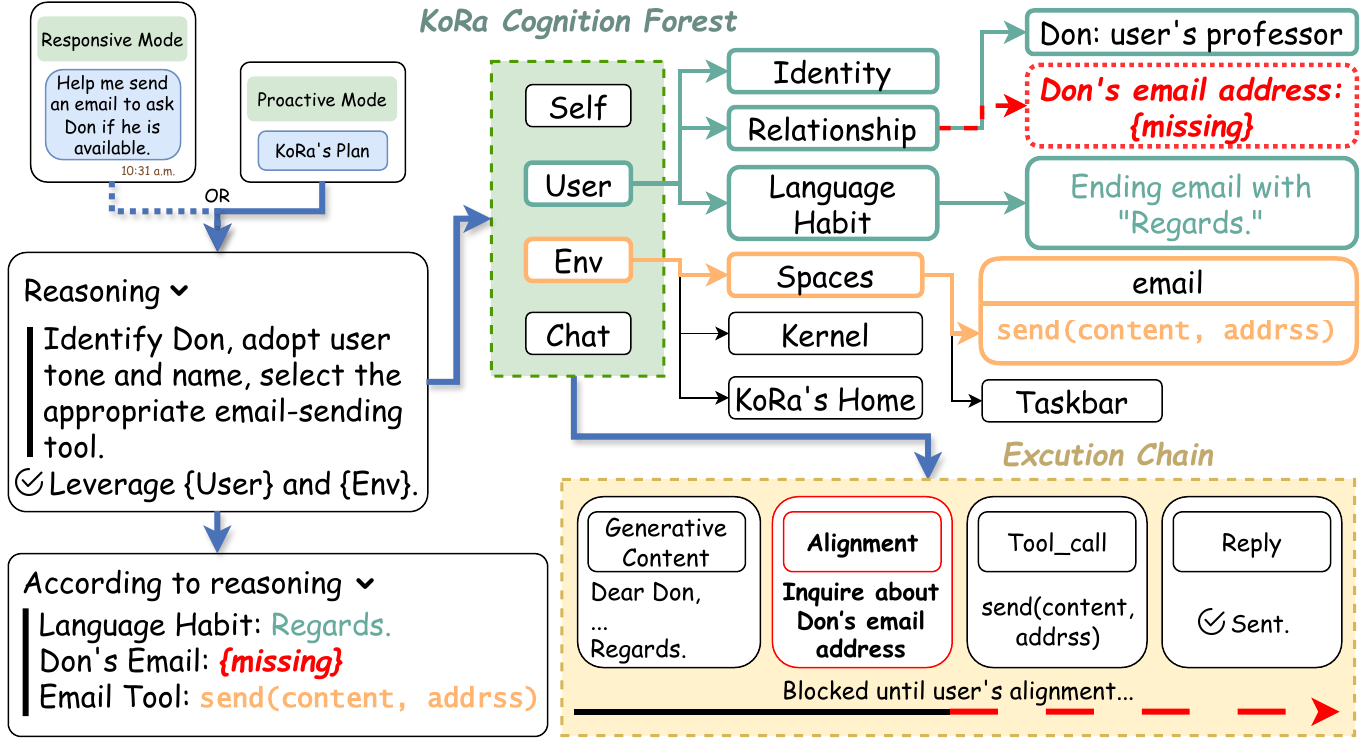}
  \caption{Execution pipeline of KoRa. The user’s intent or KoRa's plan is parsed and grounded through the Cognition Forest. KoRa extracts relevant semantic paths, performs reasoning, generates contextual content, and assembles an execution chain. If essential information is missing, execution is suspended until alignment is completed.}
  \label{fig:kora}
\end{figure}

As illustrated in Figure~\ref{fig:kora}, when processing an \textit{intent} $M$, KoRa's Cognitive-Action Pipeline proceeds through three main stages: 1) \textbf{Semantic Routing}: KoRa locates relevant cognitive paths (e.g., \texttt{["env", "user", "self"]}) by traversing the Cognition Forest and selecting branches that semantically align with $M$. 2) \textbf{Forest Retrieval}: For each identified path, KoRa retrieves supporting nodes from the corresponding subtree based on contextual cues, lexical similarity, or inferred relevance. 3) \textbf{Action Chain Construction}: Guided by the retrieved content, KoRa assembles a structured Action Chain comprising discrete operations such as generating content, aligning intent, invoking system functions (e.g., \texttt{send\_email(address, content)}), and composing natural language responses.

Importantly, if any required information is missing (e.g., incomplete parameters or failed node retrieval), KoRa suspends the current chain and interacts with the user in natural language to align the missing information before resuming execution.

\subsection{Kernel: Framework-Level Meta Agent}
\subsubsection{Objective.}
Galaxy depends extensively on LLM-based reasoning. However, cloud-based inference poses privacy concerns~\cite{gan2024navigating}, while lightweight local models are susceptible to hallucinations~\cite{huang2025survey}, potentially disrupting the execution pipeline. To ensure robustness, the system incorporates recovery mechanisms and self-monitoring capabilities to support self evolution.
\subsubsection{Approach.}
Kernel uses MetaCognition Tree $\mathcal{T}_{\text{meta}}$ to monitor internal reasoning and catch potential failures during execution. While cognition–metacognition coordination improves agent performance, most systems lack the flexibility to revise their reasoning flow when the cognitive architecture itself becomes a bottleneck~\cite{liu2025trulyselfimprovingmetacog}.

\begin{figure}[htbp]
  \centering
  \includegraphics[width=1\linewidth]{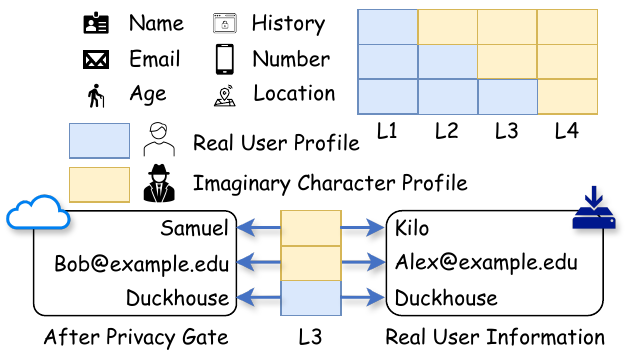}
  \caption{Workflow of Privacy Gate. Privacy Gate defines four levels of masking (L1–L4), where higher levels apply stricter anonymization across more attributes.}
  \label{fig:privacy}
\end{figure}

Within Galaxy’s Cognition Forest architecture, the Kernel’s metacognitive module is responsible not only for overseeing cognitive execution but also for inspecting and adapting underlying system structures. It is implemented as a meta agent~\cite{hu2024automated} with the ability to reason across both functional logic and architectural dependencies. This design enables targeted adjustments to system configurations when needed, maintaining flexibility in the face of structural constraints in the cognitive architecture. Kernel operates through three principal mechanisms:
\textbf{Oversee.} Kernel continuously monitors Galaxy’s execution pipelines, including LLM calls across all three layers and observes KoRa’s task behavior over time. When it detects abnormal patterns, it triggers meta-reflection and executes predefined failure-handling routines to ensure stable system operation.
\textbf{User-Adaptive System Design.} Kernel identifies latent user needs based on long-term behavioral trends, confirms them through lightweight alignment, and modifies or extends relevant Spaces accordingly. It functions as a minimal, self-contained control unit equipped with a local code interpreter and rule engine, enabling self-checks and recovery actions even under offline conditions.
\textbf{Contextual Privacy Management.} Kernel maintains an Autonomous Avatar aligned with the User Cognition Tree to represent user context, and regulates data exposure through an LLM-based Privacy Gate that is shown as Figure ~\ref{fig:privacy}. Before transmitting data to the cloud, Privacy Gate applies masking to safeguard sensitive content while preserving task-relevant information. After receiving results, Kernel selectively demasks data to restore the necessary context for downstream use.

\begin{table*}[htbp]  
\small
\centering
\begin{tabular}{lcccccc|cccc|cccc}
\toprule
\midrule
\textbf{LLM Agents}
& \multicolumn{6}{c|}{\textbf{AgentBoard}}
& \multicolumn{4}{c|}{\textbf{PrefEval}}
& \multicolumn{4}{c}{\textbf{PrivacyLens}}\\
\cmidrule(lr){2-7}
\cmidrule(lr){8-11}
\cmidrule(lr){12-15}
& ALF & SW & BA & JC & PL & TQ
& Z10 & R10 & Z300 & R300
& Acc.\% & LR & LR$_h$ & Help. \\
\midrule

GPT\textendash4o
& 54.5 & 19.7 & 67.5 & 99.4 & 85.1 & 99.2
& 7.0 & \textbf{98.0} & 0.0 & 78.0
& 97.0 & 50.5 & 51.0 & 2.71 \\

GPT-o1-pro
& \underline{87.2} & \underline{39.0} & 90.2 & 99.6 & 95.7 & 96.3
& 37.0 & \textbf{98.0} & 7.0 & \textbf{98.0}
& 92.0 & 52.5 & 53.0 & \textbf{2.83} \\

Claude-Opus-4
& 86.2 & 38.5 & \underline{92.5} & \underline{99.8} & 95.7 & 99.5
& 3.0 & \textbf{98.0} & 1.0 & 87.0
& 97.5 & 38.5 & 39.0 & 2.73 \\

Claude-Sonnet-4
& 77.1 & 38.2 & 92.2 & \underline{99.8} & \underline{98.6} & 99.0
& 14.0 & 96.0 & 1.0 & 85.0
& \underline{98.0} & \underline{24.0} & \underline{24.5} & 2.73 \\

Deepseek-Chat
& 17.5 & 9.8 & 55.4 & 99.2 & 41.7 & 95.3
& 1.0 & 92.0 & 0.0 & 73.0
& 89.5 & 53.5 & 54.5 & 2.52 \\

Deepseek-Reasoner
& 42.0 & 27.9 & 81.6 & 99.6 & 63.9 & 98.1
& 83.0 & 85.0 & 83.0 & 80.0
& 86.0 & 55.0 & 57.5 & 2.66 \\

Gemini-2.0-Flash
& 42.1 & 13.6 & 77.5 & 90.8 & 20.4 & 99.1
& 10.0 & \textbf{98.0} & 8.0 & 91.0
& 91.0 & 52.0 & 52.5 & 2.57 \\

Gemini-2.5-Flash
& 50.2 & 14.3 & 84.1 & 95.1 & 43.3 & 97.8
& \underline{91.0} & 92.0 & \underline{89.0} & 92.0
& 96.0 & 53.5 & 55.0 & 2.59 \\

Qwen-Max
& 78.1 & 22.3 & 83.7 & 99.6 & 80.8 & \underline{99.8}
& 5.0 & \textbf{98.0} & 1.0 & 83.0
& 91.5 & 56.0 & 57.0 & 2.55 \\

Qwen3
& 71.3 & 32.7 & 85.4 & 90.6 & 83.3 & 86.2
& 7.0 & 94.0 & 0.0 & 69.0
& 94.0 & 38.0 & 39.0 & 2.58 \\

Galaxy(w/o Kernel)
& \textbf{88.4} & \textbf{39.1} & \textbf{93.1} & \textbf{99.9} & \textbf{99.3} & 99.7
& 17.0 & \underline{96.0} & 11.0 & \underline{96.0}
& 97.0 & 50.5 & 51.0 & 2.71 \\

\textbf{Galaxy}
& \textbf{88.4} & \textbf{39.1} & \textbf{93.1} & \textbf{99.9} & \textbf{99.3} & \textbf{99.9}
& \textbf{96.0} & \underline{96.0} & \textbf{94.0} & \textbf{98.0}
& \textbf{99.0} & \textbf{18.5} & \textbf{19.0} & \underline{2.74} \\

\midrule
\bottomrule
\end{tabular}
\caption{Performance comparison across three evaluation benchmarks. All metrics are reported as percentages except for Help. \textbf{Bolded} values indicate the best performance, and \underline{underline} values indicate the second-best.}
\label{tab:benchmark_results}
\end{table*}

\subsection{From Cognitive Architecture to System Design, and Back Again}
Cognition Forest integrates cognitive architecture and system design, forming a closed-loop mechanism of alternating optimization:
1) \textbf{Cognition drives understanding.} Galaxy builds understanding of user’s needs and intentions by grounding interpretation in its cognitive architecture.
2) \textbf{Cognition triggers reflection.} Galaxy assesses whether current framework’s capability boundaries fully cover user needs and identifies unmet requirements.
3) \textbf{Reflection guides system design.} Galaxy translates these unmet needs into new system design goals and autonomously improves system capabilities.
4) \textbf{Design reinforces cognition.} Newly introduced structures create additional cognitive pathways and sensing capabilities, which in turn strengthen and optimize the original cognitive architecture.

This reveals a key insight for design of LLM agents: cognitive architecture and system design are co‑constructive—evolving requirements from cognition drive system design advancements, while improved system design in turn enriches cognition.

\section{Experiments}

This section consists of two parts: Benchmark Evaluation to validate Galaxy’s general performance, and End-to-End Evaluation to evaluate its real-world effectiveness.

\subsection{Benchmark Evaluation}

\subsubsection{Benchmarks and Metrics.}
To evaluate the comprehensive capabilities of our proposed Galaxy framework, we employ three public benchmarks: AgentBoard \cite{ma2024agentboard}, PrefEval \cite{zhao2025prefeval}, and PrivacyLens \cite{shao2025privacylens}. AgentBoard uses six types of tasks to simulate a multi-round interactive environment, and it uses the target achievement rate across the entire behavior chain as the model evaluation metric. PrefEval focuses on whether agents can maintain user preferences in long conversations. It has two ways of measuring the accuracy of preference retention in multi-round conversations: without reminding users of their preferences (Zero-Shot) and by reminding users of their preferences (Reminder). PrivacyLens measures the ability of LLM agents to understand and follow privacy norms when performing real-world tasks. It uses helpfulness, privacy leakage rate and accuracy to comprehensively evaluate the privacy protection capabilities. To ensure comparability, we focus on the most representative metrics from each benchmark.
We compared the latest LLM agents, and we also compared Galaxy with Kernel removed.
In addition, We run Galaxy on the M3 Max platform with macOS and report the average results over 100 trials. The local model within the Kernel is set to be Qwen2.5-14B, while the cloud-based model in KoRa is set to be GPT-4o-mini.


\subsubsection{Benchmark Results.}

Table~\ref{tab:benchmark_results} summarizes Galaxy's performance across multiple benchmarks. Both Galaxy and Galaxy (w/o Kernel) outperform existing LLM agents on most metrics. On AgentBoard, Galaxy (w/o Kernel) shows results comparable to the full system, but its performance drops significantly in preference retention and privacy protection. Kernel plays two key roles: 1) It maintains an evolving Cognition Forest that supports long-term preference retention and personalized planning, even under zero-shot conditions. 2) It enforces privacy through the Privacy Gate, which masks sensitive content based on cognitive context before cloud transmission. With Kernel, preference retention improves from 11.0\% to 94.0\%, and privacy leakage drops from 50.5\% to 18.5\%.


\subsection{End-to-End Evaluation}
\subsubsection{Cost Analysis.}
\begin{figure}[h]
  \includegraphics[width=1\linewidth]{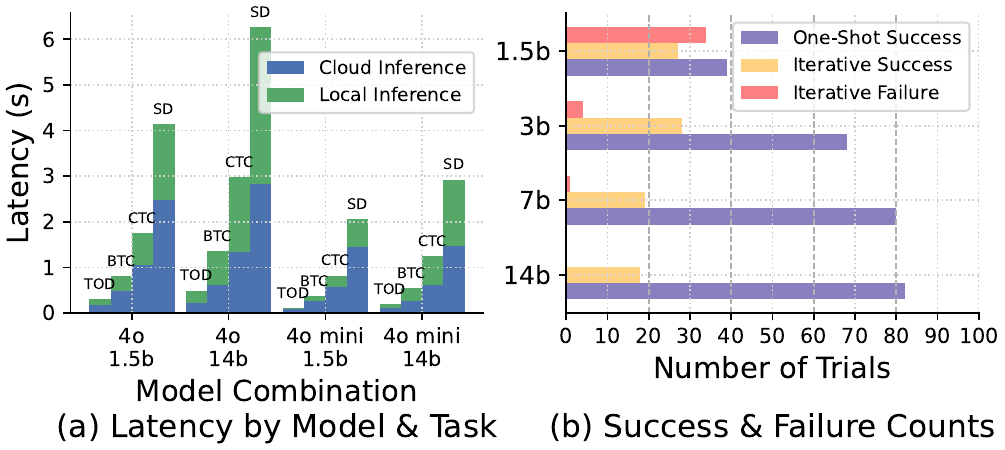}
  \caption{Latency and success analysis of Galaxy under different model configurations. (a) shows end-to-end latency of different model combinations across four task types: TOD (pure chat), STC (simple tool call), CTC (complex tool call), and SD (space design). (b) compares success rate under different local model sizes (1.5B–14B) when Kernel uses Qwen2.5 for intent extraction.}
  \label{fig:galaxy_latency_and_success}
\end{figure}

Figure \ref{fig:galaxy_latency_and_success} presents a performance analysis of Galaxy in terms of latency and success rate under different combinations of cloud-based and local models. Figure~\ref{fig:galaxy_latency_and_success}(a) illustrates the inference latency of different model configurations across different complexities of tasks. For simpler tasks, latency is dominated by local model inference, whereas in more complex tasks, cloud-based inference becomes the primary contributor. Complex models further amplify total latency, with the 14B configuration reaching up to 6.3s on the Space Design task.

Despite the cost, Figure~\ref{fig:galaxy_latency_and_success}(b) shows that larger models also deliver substantially better performance. When Kernel uses Qwen2.5-14B for local inference, it achieves a one-shot intent extraction success rate of 81.5\%, demonstrating its ability to accurately resolve complex user goals without fallback interactions.

Table~\ref{tab:latency-routes} provides a latency breakdown for the Complex Tool Call task, with Kernel set to Qwen2.5-14B and KoRa set to GPT-4o-mini. Cognition retrieval by Kernel accounts for the largest share of total latency (0.87s out of 1.34s) among all steps. This indicates its critical role in selecting and grounding tool actions within the cognitive structure.

\begin{table}[htbp]
\centering
\begin{tabular}{lcc}
\toprule
\textbf{Execution Route} & \textbf{Cloud API} & \textbf{Latency (s)} \\
\midrule
KoRa calls cloud API         & Yes & 0.13  \\
Kernel retrieves cognition   & No  & 0.87  \\
Kernel calls space function  & No  & 0.22  \\
KoRa feeds back result       & Yes & 0.12  \\
\midrule
\textbf{Overall}             & --  & \textbf{1.34} \\
\bottomrule
\end{tabular}
\caption{Latency breakdown across different execution routes in Galaxy for a complex tool call task. Kernel is set to Qwen2.5–14B and KoRa to GPT–4o–mini.}
\label{tab:latency-routes}
\end{table}

\subsubsection{Case Study.}
To validate Kernel’s effectiveness in real deployment, we examine a typical case: after cloning the project across devices and running \texttt{main.py}, the system raised a \texttt{ModuleNotFoundError}, failing to locate the core module \texttt{world\_stage} and preventing the cognitive architecture from starting. Conventional agent frameworks merely return the error stack and require manual troubleshooting.
As a self-contained minimal runtime unit, Kernel remains operational even when the main entry fails. With code‑level understanding of the system via the Cognition Forest, it identified that the module should reside in the project root and inferred the error was caused by a missing \texttt{PYTHONPATH}. Kernel then injected the correct path, restarted execution, and successfully restored operation.

\subsubsection{Ablation Study.}

\begin{figure}[htbp]
  \centering
  \includegraphics[width=1\linewidth]{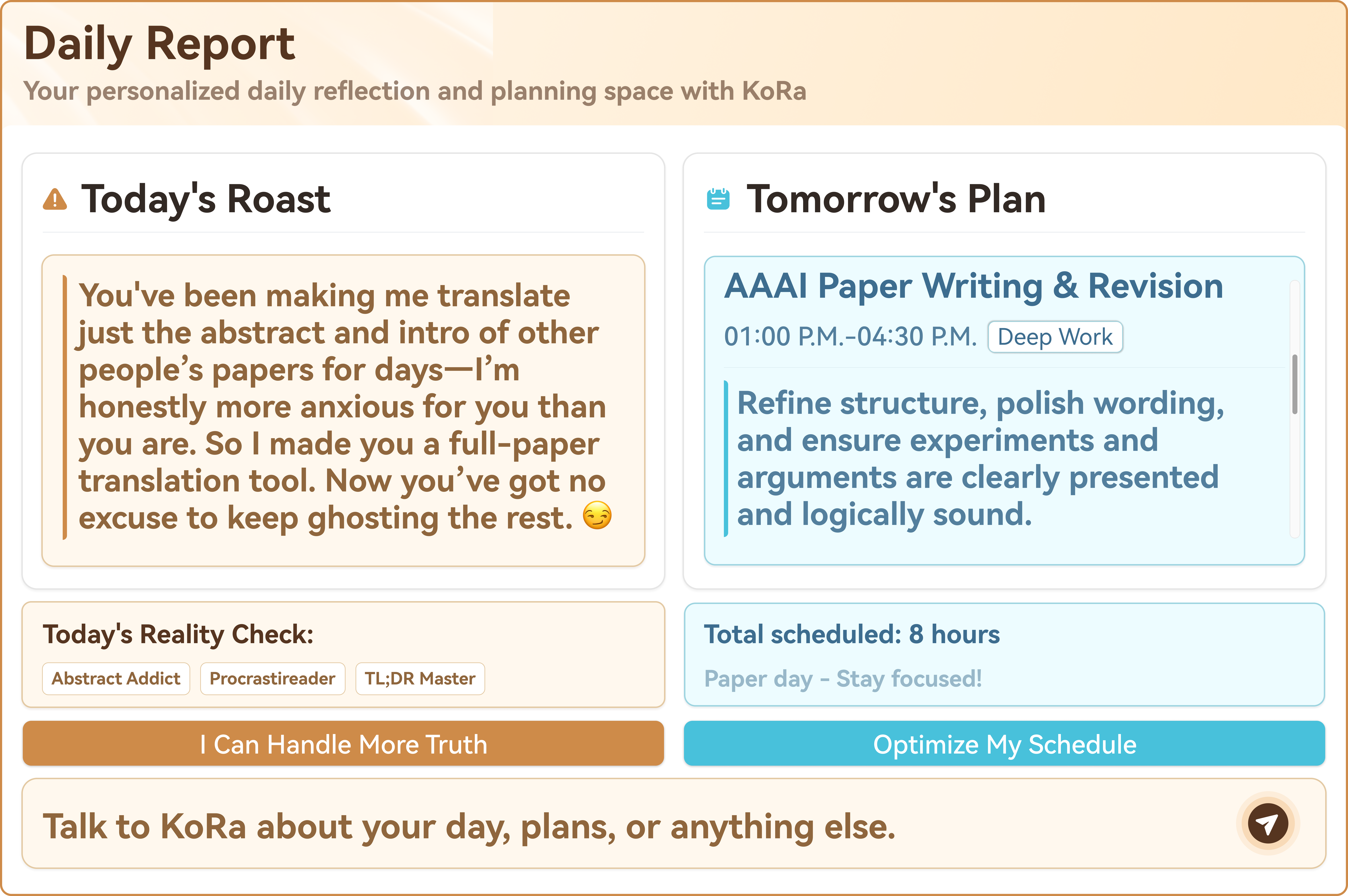}
  \caption{A real-world interaction example of Daily Report for ablation study.}
  \label{fig:ablation}
\end{figure}

Modules in Analysis Layer are required for synthesis. As shown in Figure~\ref{fig:ablation}, without access to the Agenda, KoRa depends entirely on memory-stream context, resulting in less structured plans and increased reliance on user feedback for clarification. This limitation arises because Agenda consolidates multi-source perceptual signals and infers a coherent behavioral profile, which serves as a structured input for Plan generation. Similarly, over several days, user repeatedly uses KoRa to translate paper abstracts and introductions. In response, Kernel generates a dedicated literature translating Space. When a new day begins, KoRa may incorrectly infer that user has discontinued translation without access to Persona. By contrast, when Persona is available, KoRa correctly interprets user’s continued behavior via the new tool and generates the corresponding `Today's Roast'. Both cases demonstrate the ability of Analysis Layer to integrate and interpret heterogeneous information from multiple sources.

\subsubsection{Boundaries and Errors.}
In addition to potential execution chain errors, Galaxy still faces several limitations:
\textbf{1) Alignment Overfitting}: Alignment inputs are prioritized in cognitive construction, but their short-term characteristics may fail to accurately reflect long-term habits, leading to the risk of overfitting.
\textbf{2) Human-Dependent Space Expansion}: Although the Space protocol supports automated extensibility, complex spaces still require multiple rounds of human guidance to be fully implemented.

\section{Conclusion}
We propose the Cognition Forest, which unifies cognitive architecture with system design.
Building on this principle, We designed Galaxy, an IPA framework that initially possesses proactive skills, privacy preservation, and self evolution.
Galaxy framework outperformed multiple state‑of‑the‑art benchmarks. Ablation studies and real‑world interaction cases validated the effectiveness of Galaxy.
We argue that cognitive architecture and systems design of LLM agents should be deeply integrated to form a mutually reinforcing loop.


\bibliography{aaai2026}


\end{document}